\documentclass{article}

\usepackage{microtype}
\usepackage{graphicx}
\usepackage{subcaption}
\usepackage{booktabs} 

\usepackage{hyperref}
\usepackage{xspace}

\usepackage{url}
\usepackage{bbm}

\usepackage[preprint]{icml2026}

\usepackage{amsmath}
\usepackage{amssymb}
\usepackage{mathtools}
\usepackage{amsthm}

\usepackage{siunitx}

\usepackage{enumitem}
\newcommand{\vpara}[1]{\vspace{0.04in}\noindent\textbf{#1}\xspace}

\usepackage[capitalize,noabbrev]{cleveref}

\theoremstyle{plain}

\theoremstyle{definition}

\theoremstyle{remark}

\usepackage{color}
\usepackage{colortbl}
\usepackage{tcolorbox}
\usepackage[dvipsnames]{xcolor}

\definecolor{darkblue}{rgb}{0, 0, 0.5}
\hypersetup{colorlinks=true, citecolor=darkblue, linkcolor=darkblue, urlcolor=darkblue}

\usepackage[textsize=tiny]{todonotes}

\icmltitlerunning{EvoClinician: A Self-Evolving Agent for Multi-Turn Medical Diagnosis via Test-Time Evolutionary Learning
}

\begin{document}

\twocolumn[
  \icmltitle{EvoClinician: A Self-Evolving Agent for Multi-Turn Medical Diagnosis via Test-Time Evolutionary Learning}



  \icmlsetsymbol{intern}{*}
  \icmlsetsymbol{cr}{$\flat$}

  \begin{icmlauthorlist}
    \icmlauthor{Yufei He}{comp,nus,intern}
    \icmlauthor{Juncheng Liu}{comp}
    \icmlauthor{Zhiyuan Hu}{nus}
    \icmlauthor{Yulin Chen}{nus}
    \icmlauthor{Yue Liu}{nus}
    \icmlauthor{Yuan Sui}{nus}
    \icmlauthor{Yibo Li}{nus}
    \icmlauthor{Nuo Chen}{nus}
    \icmlauthor{Jun Hu}{nus}
    \icmlauthor{Bryan Hooi}{nus}
    \icmlauthor{Xinxing Xu}{comp,cr}
    \icmlauthor{Jiang Bian}{comp}
  \end{icmlauthorlist}
  \icmlaffiliation{comp}{Microsoft Research}
  \icmlaffiliation{nus}{National University of Singapore}
  \icmlcorrespondingauthor{Xinxing Xu}{xinxingxu@microsoft.com}

  \icmlkeywords{Machine Learning, ICML}

  \vskip 0.3in
]



\printAffiliationsAndNotice{\icmlEqualContribution}

\begin{abstract}
Prevailing medical AI operates on an unrealistic ``one-shot'' model, diagnosing from a complete patient file.
However, real-world diagnosis is an iterative inquiry where Clinicians sequentially ask questions and order tests to strategically gather information while managing cost and time.
To address this, we first propose \textbf{Med-Inquire}, a new benchmark designed to evaluate an agent's ability to perform multi-turn diagnosis. 
Built upon a dataset of real-world clinical cases, Med-Inquire simulates the diagnostic process by hiding a complete patient file behind specialized \texttt{Patient} and \texttt{Examination} agents. They force the agent to proactively ask questions and order tests to gather information piece by piece.
To tackle the challenges posed by Med-Inquire, we then introduce \textbf{EvoClinician}, a self-evolving agent that learns efficient diagnostic strategies at test time. 
Its core is a ``Diagnose-Grade-Evolve'' loop: an \texttt{Actor} agent attempts a diagnosis; a \texttt{Process Grader} agent performs credit assignment by evaluating each action for both clinical yield and resource efficiency; finally, an \texttt{Evolver} agent uses this feedback to update the Actor's strategy by evolving its prompt and memory. 
Our experiments show EvoClinician outperforms continual learning baselines and other self-evolving agents like memory agents.
The code is available at \href{https://github.com/yf-he/EvoClinician}{https://github.com/yf-he/EvoClinician}




\end{abstract}
\section{Introduction}
The application of artificial intelligence in medicine holds immense promise for clinical decision-making~\citep{cabral2024clinical,nori2023capabilities,gilson2022does}. 
However, a prevailing paradigm for medical AI operates on an unrealistic ``one-shot'' model, where an agent is fed a complete patient file—with all history, findings, and lab results—to produce a diagnosis~\citep{nori2023capabilities}. 
This static, complete-information approach stands in contrast to real-world clinical practice.
A clinician’s work is not a single calculation but a dynamic and iterative process of inquiry~\citep{nori2025sequential}. 
They begin with a chief complaint and sequentially ask targeted questions, perform physical exams, and order specific tests to strategically gather information, all while judiciously managing constraints of cost and time~\citep{li2024agent,liberatore2008analytic}. 
This procedural skill of balancing diagnostic accuracy with resource efficiency is the sign of an expert clinician, yet it is a capability largely absent in today's medical AI systems~\citep{nori2025sequential}.

Existing benchmarks primarily measure final diagnostic accuracy, failing to reward or even quantify an agent's efficiency in arriving at that diagnosis~\citep{hu2024omnimedvqa,wang2024cmb,ye2024gmai}. 
To provide a more realistic measure of clinical ability, we introduce \textbf{Med-Inquire}, a benchmark built upon real-world clinical cases. It enforces a sequential diagnostic workflow by concealing the complete patient file behind specialized \texttt{Patient} and \texttt{Examination} agents. This ``gatekeeper'' design compels agents to learn the skill of proactive inquiry, gathering information piece by piece as a human clinician would.
The agent's performance is assessed by two key components. A \texttt{Judge Agent} compares the submitted diagnosis against the ground truth, assigning a score that reflects its correctness. 
Simultaneously, a \texttt{Cost Estimator} agent tracks the cumulative cost of the clinical encounter.

To this end, we introduce \textbf{EvoClinician}, a self-evolving agent that learns efficient diagnostic strategies by treating each unique case in the Med-Inquire benchmark sequentially as a distinct, test-time learning opportunity.
The agent's learning process is a ``Diagnose-Grade-Evolve'' loop, driven by an internal multi-agent architecture.

First, in the \textbf{Diagnose} phase, an \texttt{Actor} agent, guided by its current prompt and memory, attempts to diagnose a new case. It interacts with the Med-Inquire environment, asking questions and ordering tests until it decides to submit a final diagnosis. The full transcript of this interaction, including every action and its outcome, is logged.

Next, the \textbf{Grade} phase solves the difficult credit assignment problem inherent in long-horizon tasks. Instead of relying on a single, sparse reward from the final judgment, a \texttt{Process Grader} agent performs a detailed post-session review. It analyzes the full transcript, the final judgment, and the cost report to generate a dense feedback signal. It assigns a qualitative grade (e.g., \textit{`HIGH\_YIELD'}, \textit{`LOW\_YIELD'}, \textit{`INEFFICIENT'}, \textit{`CRITICAL\_ERROR'}) and a rationale to \textit{each individual action} taken by the \texttt{Actor}. 
This transforms the evaluation from a simple trajectory-level score into an action-level critique.

Finally, in the \textbf{Evolve} phase, this granular feedback is used to create lasting improvements. An \texttt{Evolver} agent takes the action-level critiques from the \texttt{Process Grader} and performs a gradient-free update on the \texttt{Actor}'s core strategy. This update is twofold, modifying both the agent's high-level strategic guidance (its prompt) and its specific experiential knowledge (its memory).

For \textbf{Prompt Evolution}, the \texttt{Evolver} translates the action-level grades into permanent changes in the \texttt{Actor}'s prompt. 
High-yield actions are abstracted into generalizable instructions for future use, teaching the agent what to do. For example, a successful early question might become a new rule: ``If a patient has a scalp lump, always ask if it has been present since birth.'' 
Conversely, inefficient or erroneous actions are used to teach the agent what not to do. 
For \textbf{Memory Evolution}, it captures both positive and negative experiences at the action level. The \texttt{Evolver} processes significant action from the session. It creates a discrete memory entry that includes: the clinical context (the state) before the action, the action itself, the resulting new information (the outcome), and the grade it received. 
This creates a rich repository of both effective and bad moves. In subsequent cases, when the \texttt{Actor} faces a similar clinical context, it can retrieve relevant memories. 


More generally, the same methodology applies to any domain where an agent must plan, use tools, and make decisions under limits (time, money, risk, or compute), and where feedback is not immediately and delayed across many steps. For example, in customer service the agent must choose among clarification, policy lookup, account actions, and escalation while balancing resolution quality and handle time. In financial management, an agent may choose data pulls, checks, and transaction proposals while controlling risk and fees, with constraints such as budgets and compliance rules. In each case, one can build an interactive environment with domain-specific tools and constraints, define a task-level score and a resource report, and then use action-level critiques to update reusable rules and memory.

We summarize our contributions as follows:
\begin{itemize}[leftmargin=*,itemsep=0pt,parsep=0.2em,topsep=0.3em,partopsep=0.3em]
\item We introduce \textbf{Med-Inquire}, an interactive benchmark for multi-turn medical diagnosis that hides the full case file behind \texttt{Patient} and \texttt{Examination} agents and evaluates both diagnosis grade and resource cost.
\item We propose \textbf{EvoClinician}, a test-time learning agent with a Diagnose-Grade-Evolve loop that uses action-level grading to update prompt and memory between cases.
\item We compare against static prompting, memory agents, prompt optimization methods, and prior self-evolving agents, and we show that our approach improves outcome quality while reducing resource use.
\end{itemize}
\section{Related Work}
\subsection{Medical Agents}
Most medical language model evaluations are still based on static exam questions, which bundle history and examination results into one input. 
This style has been used in many studies of medical question answering and diagnostic reasoning, including work reporting strong performance on medical challenge problems and licensing style tests \citep{nori2023capabilities, singhal2022large, gilson2022does, cabral2024clinical}. 
Related interactive medical work studies question-asking as a first-class decision problem\citep{gorry1968experience,horvitz1988decision}. MedIQ \citep{li2024mediq} evaluates question-asking LLMs for interactive clinical reasoning. 
SDBench \citep{nori2025sequential}, built from NEJM clinicopathological conference cases, evaluates both diagnostic correctness and cost. 


There is also a growing body of work on multi-agent clinical simulation, where doctor, patient, and tool roles interact in a simulated setting. 
These systems are often used for studying dialogue quality, coordination, and training data generation \citep{li2024agent, almansoori2025self, chen2025mdteamgpt}. AgentClinic \citep{schmidgall2024agentclinic} is another related direction that benchmarks multimodal agents in simulated clinical environments. Compared with these simulation systems, Med-Inquire is designed for measuring sequential diagnostic policies with a reproducible cost model and a diagnosis grading rubric.

\begin{figure*}[t]
    \centering
    \includegraphics[width=1.0\textwidth]{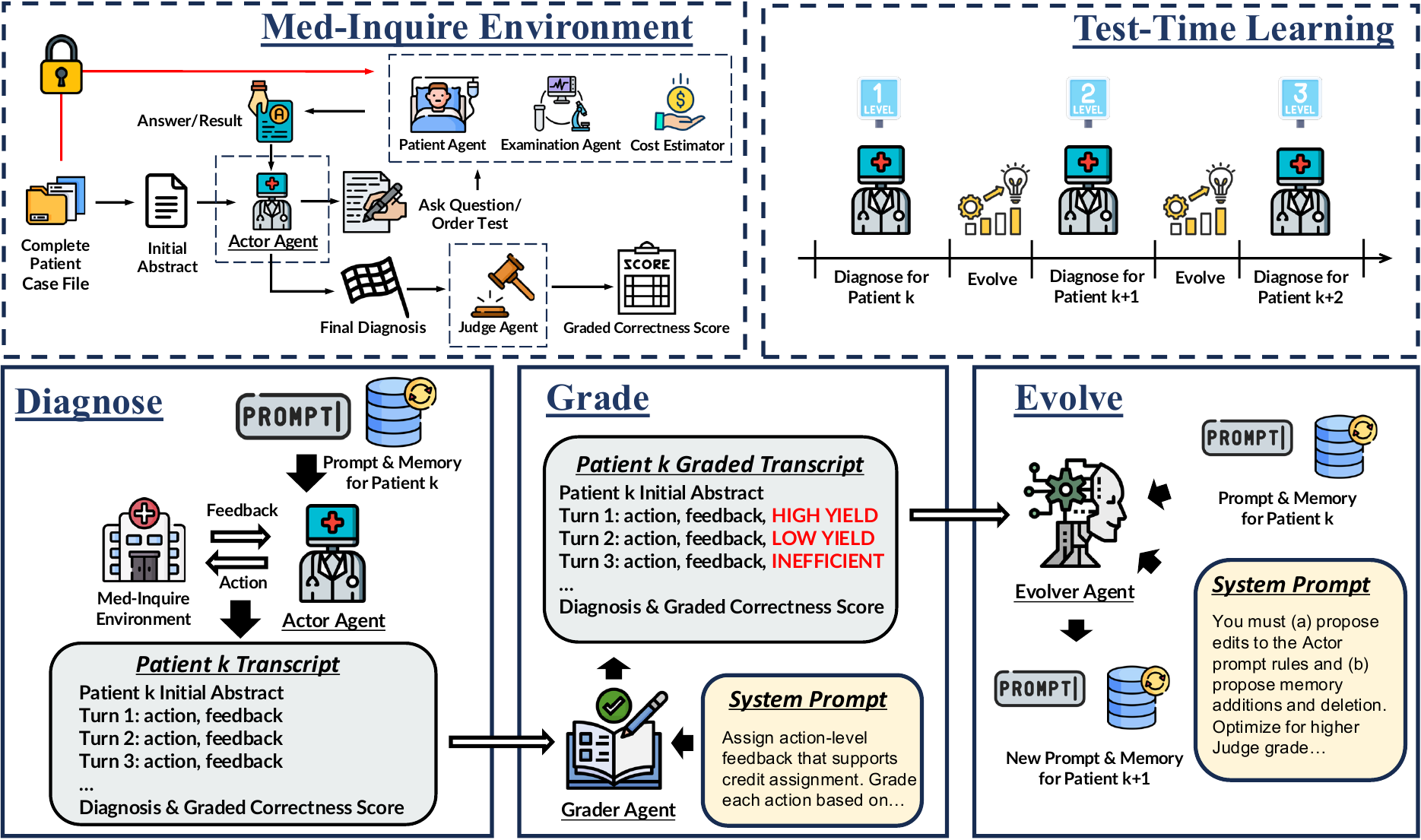} 
    \caption{\textbf{EvoClinician architecture and test-time learning loop.} The Actor interacts with the Med-Inquire environment through \texttt{AskQuestion} and \texttt{OrderTest}, receiving responses from the Patient and Examination agents, while the Cost Estimator tracks resource use. After \texttt{SubmitDiagnosis}, the Judge assigns a graded correctness score. The Process Grader then reviews the full transcript and assigns action-level labels (for example, HIGH YIELD, LOW YIELD, INEFFICIENT) for credit assignment. Finally, the Evolver uses these labels, the final score, and the cost report to update the Actor's prompt rules and external memory before the next episode, forming a repeated Diagnose--Grade--Evolve loop over a stream of cases at test time.
    }
    \label{fig:architecture}
    \vspace{-6mm}
\end{figure*}

\subsection{Self-Evolving Agents}
Self-evolving agents update their behavior based on experience, rather than running with a fixed prompt and no persistent state~\citep{wang2024survey,xi2025rise,he2025evaluating,li2026just,hu2026rewarding,chen2025mlr}. 
A common mechanism is long-term memory that stores past interactions and retrieves relevant items for new tasks. Retrieval-augmented generation (RAG) provides a general template for adding external memory to language model inference \citep{lewis2020retrieval,he2025enabling,he2025unigraph,he2025unigraph2,he2024generalizing}. More agent-specific memory systems include Mem0, which maintains memory with add, merge, and delete operations, and trajectory-based methods that turn experience into reusable guidance \citep{chhikara2025mem0, zhao2024expel,liu2025guardreasoner,chen2025can,chen2025robustness}. 
Workflow-level memory stores reusable action plans rather than raw text, aiming to improve multi-step behavior \citep{wang2024agent}. These methods often help reuse information, but they do not always change the agent policy in a stable way, because the main policy still comes from a fixed prompt.

Another line of work updates prompts or rules. Prompt optimization can be cast as search over instruction text, including evolutionary prompt methods such as PromptBreeder\citep{fernando2023promptbreeder}, and other automatic prompt optimization frameworks \citep{pryzant2023automatic, yang2023large}. Some approaches use language feedback as an update signal for prompts or intermediate reasoning steps, including TextGrad \citep{yuksekgonul2024textgrad}. Reflection-based methods store critiques of failures and use them to guide later attempts, for example Reflexion and Self-Refine \citep{shinn2023reflexion, madaan2023self}. These methods give denser guidance than a single final score, but they often do not perform action-level credit assignment for long trajectories with costs~\citep{he2025evotest}.

Reinforcement learning methods for agents address sequential decision making more directly, but the learning signal is often sparse and hard to assign to individual actions~\citep{wang2025safety,liu2025efficient,sui2025meta,gao2025flowreasoner,sui2024fidelis,sui2024can}. Recent work studies multi-turn reinforcement learning for LLM agents and strategy learning from interaction, including RAGEN \citep{wang2025ragen} and DYSTIL \citep{wang2025dystil}. Separate work on process supervision and step verification shows that intermediate feedback can improve long-horizon reasoning, including Let’s Verify Step by Step \citep{lightman2023let} and Math-Shepherd \citep{wang2024math}. EvoClinician adopts the same general idea of denser feedback, but applies it to clinical actions (questions and tests) and adds resource efficiency to the grading signal. 

\section{Benchmark: Med-Inquire}
\label{sec:benchmark}


\subsection{Task Definition}
Med-Inquire is an interactive diagnosis task. Each episode corresponds to one real-world clinical case with a hidden full patient file. 
The agent starts with only an initial problem statement (for example, a chief complaint and basic demographics when available). The agent then takes actions to gather more information. At any time, the agent may stop and submit a final diagnosis.

We model the interaction as a partially observable Markov decision process (POMDP). The hidden state is the complete case file. The agent observes only the information returned by the environment in response to its actions. This structure matches the key limitation of clinical care: the clinician does not observe the full state and must choose what to ask or order next.

\subsection{Environment Design and Information Gatekeeping}
Med-Inquire hides the case file behind two environment components that we implement as constrained agents.

The \textbf{Patient agent} answers questions in the style of a real patient. It responds to queries about symptoms, medical history, and related context (for example, onset time, triggers, and prior conditions) using only information recorded in the hidden case file. If the case file does not contain the requested detail, it generates a medically reasonable patient-style answer that is consistent with the ground-truth diagnosis and with information already revealed in the dialogue, rather than returning an ``unknown/not documented'' response. 

The \textbf{Examination agent} returns results for physical examination findings and diagnostic tests. Tests include both low-cost screening tests and higher-cost confirmatory tests, when those are represented in the case file. If an agent orders a test that is not present in the case file, the Examination agent returns ``not available''.


\subsection{Action Space and Interaction Protocol}
At each turn $t$, the agent chooses one action $a_t$ from three types.

First, \textbf{AskQuestion}: a free-form question to the Patient agent. The environment returns a text answer grounded in the case file.

Second, \textbf{OrderTest}: a request to the Examination agent for a named examination maneuver or diagnostic test. The environment returns the recorded test result, if present, and otherwise returns a null response as described above.

Third, \textbf{SubmitDiagnosis}: the agent submits a final diagnosis. This action ends the episode.

To make evaluation reproducible, we set a fixed maximum number of turns $T_{\max}$. We log the full transcript: all actions, all environment responses, and the final submission.

\subsection{Cost Model}
\label{sec:cost}
Med-Inquire assigns a non-negative cost to each action to reflect resource use. Let $c(a_t)$ be the cost of action $a_t$. The total encounter cost is
\begin{equation}
C = \sum_{t=1}^{T} c(a_t),
\end{equation}
where $T$ is the number of turns until the episode ends.

Costs include at least (i) a base per-turn interaction cost (time and clinician effort) and (ii) a test-specific cost for each ordered test. In our implementation, the cost schedule is provided as a table that maps test names to costs, along with a default cost for free-form questioning. The benchmark reports costs in the same unit across all cases (normalized cost units), because exact billing varies by setting and country. Users who want a different schedule can swap the cost table without changing the rest of the benchmark, as long as the mapping is reported.

\subsection{Diagnosis Judging}
After \texttt{SubmitDiagnosis}, a \textbf{Judge agent} compares the submitted diagnosis against the ground-truth diagnosis recorded in the hidden case file. The Judge agent outputs a graded correctness score $S \in [0,100]$, where higher values mean closer agreement with the ground truth.
The Judge agent uses a single graded mode. It then evaluates clinical agreement between the submission and the ground truth using the rubric in Appendix~\ref{sec:judge}. 


\subsection{Benchmark Metrics}
\label{sec:metrics}
Med-Inquire evaluates an agent along three axes: diagnostic grade, interaction length, and resource cost. For each episode $i$, the benchmark records (i) the Judge grade $S_i \in [0,100]$, (ii) the number of turns $T_i$, and (iii) the total encounter cost $C_i$.
The \textbf{diagnostic grade} $S_i$ measures how closely the submitted diagnosis matches the ground-truth diagnosis under the rubric in Appendix~\ref{sec:judge}. 
The \textbf{number of turns} $T_i$ counts all actions taken until the agent submits a diagnosis. 
The \textbf{cost} $C_i$ is the sum of per-action costs.

\section{Method: EvoClinician}
\label{sec:method}

\subsection{Test-Time Learning Setting}
\label{sec:ttl}
We use the term \emph{test-time learning} (TTL)~\citep{he2025enabling,he2025evotest} to indicate any method that updates its internal state during evaluation, after deployment of the base model, using only information available from the test-time interaction. 
In our setting, the agent faces a sequence of diagnosis episodes $\{1,\dots,N\}$ from Med-Inquire. During episode $i$, the agent can only observe (1) the initial case presentation, (2) responses from the Patient and Examination agents to its actions, (3) the final Judge score $S_i \in [0,100]$, and (4) the total cost $C_i$. The agent does not receive the hidden case file or the ground-truth diagnosis information.

We study TTL where the agent updates \emph{prompts} and \emph{external memory} between episodes. We do not update the weights of the base LLMs. 

\subsection{Overview}
\label{sec:overview}
As shown in Figure~\ref{fig:architecture}, EvoClinician is a multi-agent system that runs a repeated \textbf{Diagnose--Grade--Evolve} loop. It has three internal roles.

First, an \textbf{Actor} interacts with the Med-Inquire environment to gather information and then submits a diagnosis.

Second, a \textbf{Process Grader} reads the full transcript, the final Judge score, and the cost report, and then assigns action-level feedback. This feedback is meant to address credit assignment: it links later success or failure to earlier questions and tests.

Third, an \textbf{Evolver} updates the Actor's strategy for future episodes. The Evolver edits (i) the Actor prompt, which acts as persistent high-level guidance, and (ii) an external memory, which stores action advice with labels.

\subsection{Actor}
\label{sec:actor}
The Actor is an LLM agent with a system prompt $P$ and a memory $M$. 
At turn $t$, the Actor receives the dialogue history $h_t$ and retrieves a small set of relevant memories
\begin{equation}
\mathcal{R}_t = \mathrm{Retrieve}(M, h_t; k),
\end{equation}
where $k$ is a fixed retrieval count.

The Actor then chooses an action $a_t \in \{\texttt{AskQuestion}, \texttt{OrderTest}, \texttt{SubmitDiagnosis}\}$ and produces natural-language content for that action (the question, the test name, or the diagnosis). The environment returns an observation $o_{t+1}$ and the running cost increases by $c(a_t)$. The episode ends when the Actor chooses \texttt{SubmitDiagnosis} or reaches $T_{\max}$.

We do not assume that the Actor has access to a fixed list of allowed tests. Instead, the Actor proposes a test name in text, and the Examination agent either returns a recorded result or ``not available''. This makes the action space open-ended and places weight on planning rather than list lookup.

\subsection{Process Grader (action-level feedback)}
\label{sec:grader}
After the episode ends, the Process Grader receives the transcript
\begin{equation}
\tau_i = \{(h_t, a_t, o_{t+1})\}_{t=1}^{T_i},
\end{equation}
the final Judge score $S_i$, and the total cost $C_i$. It outputs a set of per-action labels and short rationales:
\begin{equation}
G_i = \{(g_t, r_t)\}_{t=1}^{T_i}.
\end{equation}
Each label $g_t$ is chosen from a small set such as \texttt{HIGH\_YIELD}, \texttt{LOW\_YIELD}, \texttt{INEFFICIENT}, and \texttt{CRITICAL\_ERROR}. The grader uses two criteria.

First, \emph{clinical yield}: whether the action reduced diagnostic uncertainty given the state of the dialogue at that time. For example, a targeted question that separates two likely diagnoses can be high yield.

Second, \emph{resource efficiency}: whether the same information could likely have been obtained with lower cost or fewer steps. For example, ordering a costly confirmatory test before basic screening can be graded as inefficient, even if it returns helpful information.

The Process Grader also produces a brief session summary that states why the final diagnosis achieved score $S_i$ and which earlier steps helped or harmed performance. \textbf{\textit{A theory view of the Process Grader is given in Appendix~\ref{sec:theory}.}}

\begin{table*}[t]
\caption{Results on Med-Inquire across backbone LLMs. We report the mean Judge diagnostic grade $\bar{S}$ (higher is better), mean number of turns $\bar{T}$ (lower is better), and mean total encounter cost $\bar{C}$ (lower is better) over all 915 cases, using the same turn limit, action interface, and fixed cost table for each backbone. Costs are reported in normalized units and are comparable only within this benchmark setting.}
\vspace{-1mm}
\centering
\setlength{\tabcolsep}{5.3pt}
\begin{tabular}{lccc ccc ccc ccc}
\toprule
& \multicolumn{3}{c}{\textbf{\texttt{gemini-3-pro}}} & \multicolumn{3}{c}{\textbf{\texttt{gpt-5.1}}} & \multicolumn{3}{c}{\textbf{\texttt{gpt-4o-mini}}} & \multicolumn{3}{c}{\textbf{\texttt{MedGemma-27b}}} \\
\cmidrule(lr){2-4}\cmidrule(lr){5-7}\cmidrule(lr){8-10}\cmidrule(lr){11-13}
Method
& $\bar{S}\uparrow$ & $\bar{T}\downarrow$ & $\bar{C}\downarrow$
& $\bar{S}\uparrow$ & $\bar{T}\downarrow$ & $\bar{C}\downarrow$
& $\bar{S}\uparrow$ & $\bar{T}\downarrow$ & $\bar{C}\downarrow$
& $\bar{S}\uparrow$ & $\bar{T}\downarrow$ & $\bar{C}\downarrow$ \\
\midrule

\multicolumn{13}{l}{\textbf{Static Agent}} \\
Static Prompt
& 48.2 & 9.8 & 1380
& 43.6 & 10.1 & 1410
& 38.9 & 10.4 & 1440
& 41.0 & 10.2 & 1425 \\
\midrule

\multicolumn{13}{l}{\textbf{Memory Agent}} \\
Memory
& 50.1 & 10.4 & 1510
& 45.1 & 10.7 & 1540
& 40.4 & 11.0 & 1580
& 42.8 & 10.9 & 1560 \\
RAG~\citep{lewis2020retrieval}
& 50.7 & 10.3 & 1490
& 45.7 & 10.6 & 1520
& 40.9 & 10.9 & 1550
& 43.3 & 10.8 & 1535 \\
Mem0~\citep{chhikara2025mem0}
& 51.2 & 10.1 & 1450
& 46.3 & 10.4 & 1480
& 41.7 & 10.7 & 1510
& 44.0 & 10.6 & 1495 \\
Evo-Memory~\citep{wei2025evo}
& 52.0 & 10.0 & 1435
& 47.1 & 10.3 & 1465
& 42.5 & 10.6 & 1495
& 44.8 & 10.5 & 1480 \\
\midrule

\multicolumn{13}{l}{\textbf{Prompt Optimization Agent}} \\
EvoPrompt~\citep{guo2024evoprompt}
& 53.6 & 9.7 & 1360
& 49.0 & 9.9 & 1390
& 44.1 & 10.2 & 1420
& 46.6 & 10.0 & 1405 \\
GEPA~\citep{agrawal2025gepa}
& 49.4 & 10.6 & 1540
& 44.6 & 10.9 & 1575
& 39.8 & 11.2 & 1610
& 42.2 & 11.0 & 1593 \\
\midrule

\multicolumn{13}{l}{\textbf{Evolutionary Agent}} \\
EvoTest~\cite{he2025evotest}
& 57.9 & 9.4 & 1605
& 52.5 & 9.6 & 1650
& 46.6 & 9.8 & 1690
& 49.6 & 9.7 & 1670 \\
\textbf{EvoClinician}
& \textbf{59.8} & \textbf{9.1} & \textbf{1275}
& \textbf{54.2} & \textbf{9.3} & \textbf{1310}
& \textbf{48.0} & \textbf{9.6} & \textbf{1345}
& \textbf{51.1} & \textbf{9.5} & \textbf{1328} \\
\bottomrule
\end{tabular}
\vspace{-3mm}
\label{tab:backbone_results}
\end{table*}

\subsection{Evolver (prompt and memory updates)}
\label{sec:evolver}
The Evolver takes $(\tau_i, S_i, C_i, G_i)$ and updates the Actor prompt $P$ and memory $M$ with a gradient-free procedure. 

\vpara{Prompt evolution.}
We represent the Actor prompt as a base instruction plus a list of short rules. After episode $i$, the Evolver proposes edits to this rule list. Each edit is derived from the graded transcript.

When the grader marks an action as \texttt{HIGH\_YIELD}, the Evolver may add a rule that generalizes the pattern, conditioned on the observed context. The rule form is constrained to be short and operational, for example, ``If symptom X and time course Y, ask about Z before ordering imaging.''

When the grader marks an action as \texttt{INEFFICIENT} or \texttt{CRITICAL\_ERROR}, the Evolver may add a guard rule that blocks similar actions in similar contexts, for example, ``Before ordering test T, state what decision it changes and whether a cheaper test could answer the same question.''

To reduce drift, we use two constraints. First, the prompt has a maximum rule budget $B_P$; if the budget is exceeded, the Evolver removes or merges the lowest-utility rules based on how often the grader cited them in recent episodes. Second, the Evolver does not add rules that mention case-specific identifiers; it rewrites them into general clinical features (for example, ``acute chest pain'' rather than a named patient).

\vpara{Memory evolution.}
Memory stores case-conditioned action advice. From the transcript, the Evolver selects a subset of turns $\mathcal{T}_i$ that have strong feedback (for example, \texttt{HIGH\_YIELD} or \texttt{CRITICAL\_ERROR}). For each selected turn $t \in \mathcal{T}_i$, it writes a memory entry
\begin{equation}
m = \big(\phi(h_t), a_t, o_{t+1}, g_t, r_t\big),
\end{equation}
where $\phi(h_t)$ is a compressed representation of the clinical context extracted from the dialogue history (key symptoms, time course, and already-known test results). At the start of each new turn, retrieval returns entries with similar $\phi(h_t)$, and the Actor uses their actions and grades as suggestions or warnings.
Memory has a maximum size $B_M$. When full, we delete entries with low expected reuse, measured by a combination of age and retrieval frequency. 

\begin{figure*}[t]
  \centering
  \includegraphics[width=\textwidth]{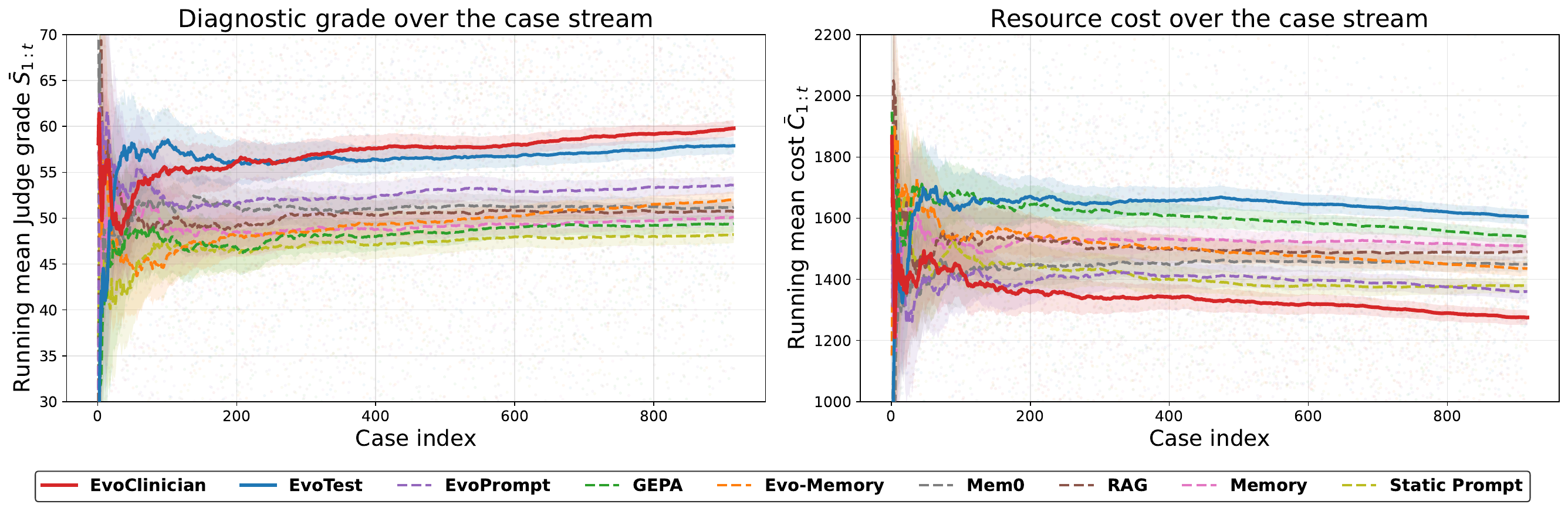}
  \vspace{-6mm}
  \caption{Running-mean learning curves on Med-Inquire over $N=915$ cases (fixed evaluation order). Left: running mean Judge grade $\bar{S}_{1:t}=\frac{1}{t}\sum_{i=1}^{t} S_i$, where $S_i\in[0,100]$ is the per-case diagnosis grade. Right: running mean cost $\bar{C}_{1:t}=\frac{1}{t}\sum_{i=1}^{t} C_i$, where $C_i$ is the per-case cumulative cost in normalized units from the fixed cost table. Faint points show per-case values, and lines show the running mean. Shaded bands show an approximate $95\%$ interval based on the running standard error, $\bar{x}_{1:t}\pm 1.96\,\mathrm{SE}_{1:t}$, computed from the first $t$ cases. }
  \label{fig:learning-curves-running}
  \vspace{-4mm}
\end{figure*}

\section{Experiments}

\subsection{Dataset}
The patient cases come from DiagnosisArena~\citep{zhu2025diagnosisarena}, which is a collection of 915 real-world medical cases taken from reports published in 10 major medical journals like \textit{The New England Journal of Medicine (NEJM)} and \textit{The Lancet}. Each case in the dataset is organized into four main parts: Case Information (the patient's background, Physical Examination (the results of check-up), Diagnostic Tests (results from labs), and the Final Diagnosis. 

\subsection{Setup}
For methods that update at test time, we evaluate cases in a fixed order and allow updates only between episodes. We run experiments with four base LLM backbones: \texttt{gemini-3-pro}~\citep{team2023gemini}, \texttt{gpt-5.1}~\citep{singh2025openai}, \texttt{gpt-4o-mini}~\citep{hurst2024gpt}, and \texttt{MedGemma-27b}~\citep{sellergren2025medgemma}. 
We run each method with the same $T_{\max}$ and the same stopping condition: the agent may submit a diagnosis at any time. If the agent reaches $T_{\max}$ without submitting, we force \texttt{SubmitDiagnosis} using the agent's current best guess.

\subsection{Baselines}
We compare three baseline categories: \textbf{static prompt}, \textbf{memory agent}, and \textbf{prompt optimization agent}. All baselines use the same base LLM, decoding settings, and setup (including the same $T_{\max}$).

The \textbf{static prompt} baseline is (1) \textbf{Static Prompt}, which uses one fixed system prompt for all episodes and stores no cross-episode state.

The \textbf{memory agent} baselines keep the prompt fixed but add long-term memory across episodes: (2) \textbf{Memory Agent}, which fills the context window with full past transcripts and drops the oldest first when reaching the token limit; (3) \textbf{RAG}~\citep{lewis2020retrieval}, which retrieves top-$k$ chunked past text from a vector index; (4) \textbf{Mem0}~\citep{chhikara2025mem0}, which extracts and consolidates memory items and retrieves them when needed; and (5) \textbf{Evo-Memory}~\citep{wei2025evo}, which update memory between episodes (experience retrieval and memory refine variants) without updating the main prompt.

The \textbf{prompt optimization agent} baselines update the system prompt for future episodes: (6) \textbf{EvoPrompt}~\citep{guo2024evoprompt}, which searches for an improved prompt on a held-out development subset and then freezes it for evaluation; and (7) \textbf{GEPA}~\citep{agrawal2025gepa}, which is a prompt optimizer that incorporates natural language reflection to learn high-level rules from trial and error. 

The \textbf{evolutionary agent} (8) \textbf{EvoTest}~\cite{he2025evotest}, which updates the agent configuration (including the prompt, memory) after each episode using transcript-level feedback.

\subsection{Evaluation}
\label{sec:evaluation}
We evaluate all methods under the test-time learning (TTL) setting in Section~\ref{sec:ttl}. Only the TTL methods are allowed to adapt across episodes (by updating prompt state and or external memory between episodes). Non-TTL baselines keep their prompt and memory fixed for all cases. 

For each episode $i$, Med-Inquire returns the Judge diagnostic grade $S_i \in [0,100]$, the number of turns $T_i$, and the total encounter cost $C_i$ (Sections~\ref{sec:metrics} and~\ref{sec:cost}). In our experiments, we report mean results over the full evaluation set of $N$ cases:
\begin{equation}
\bar{S} = \frac{1}{N}\sum_{i=1}^{N} S_i,\quad
\bar{T} = \frac{1}{N}\sum_{i=1}^{N} T_i,\quad
\bar{C} = \frac{1}{N}\sum_{i=1}^{N} C_i.
\end{equation}
These three mean values summarize diagnostic correctness ($\bar{S}$) and interaction efficiency ($\bar{T}$ and $\bar{C}$) under the same turn limit $T_{\max}$ and the same cost table for all methods.

\subsection{Results}
Table~\ref{tab:backbone_results} reports mean diagnostic grade $\bar{S}$, mean turns $\bar{T}$, and mean cost $\bar{C}$ under the same test-time learning setting (Section~\ref{sec:ttl}). In this setting, a ``self-evolving'' method changes its behavior over the case stream by updating its prompt or memory between episodes.

(1) \textbf{Test-time learning improves not only final diagnoses, but also the information-gathering policy.}
Across backbones, methods that update across episodes improve $\bar{S}$ more than methods that only retrieve past text. For example, on \texttt{gemini-3-pro}, memory methods move from 48.2 (Static Prompt) to 52.0 (Evo-Memory), while self-evolving methods reach 57.9 (EvoTest) and 59.8 (EvoClinician). The same ordering holds for \texttt{gpt-5.1}, \texttt{gpt-4o-mini}, and \texttt{MedGemma-27b} (for example, 43.6 $\rightarrow$ 47.1 $\rightarrow$ 52.5 $\rightarrow$ 54.2 on \texttt{gpt-5.1}, and 41.0 $\rightarrow$ 44.8 $\rightarrow$ 49.6 $\rightarrow$ 51.1 on \texttt{MedGemma-27b}). This is consistent with the idea that the useful signal from prior episodes includes action rules (which question to ask next, which test to avoid, and when to stop).

(2) \textbf{Cost differences are driven by which actions are chosen, not only by how many turns are used.}
Across backbones, $\bar{T}$ stays in a narrow band (about 9 to 11 turns), while $\bar{C}$ varies more. For example, on \texttt{gemini-3-pro}, EvoTest and EvoClinician use a similar number of turns (9.4 versus 9.1) but have very different costs (1605 versus 1275). The same effect appears on \texttt{gpt-5.1} (1650 versus 1310) and \texttt{MedGemma-27b} (1670 versus 1328). This suggests that cost is mainly affected by test selection (for example, ordering expensive tests early) rather than conversation length. In a multi-turn diagnosis task, efficiency therefore depends on learning which information to request and when.

(3) \textbf{The gain from self-evolving updates is smaller on weaker backbones, which suggests that adaptation needs basic reasoning and instruction-following ability.}
Absolute $\bar{S}$ decreases from stronger to weaker backbones, and the improvement from adaptation also decreases. The improvement from Static Prompt to EvoClinician is $+11.6$ on \texttt{gemini-3-pro} (48.2 $\rightarrow$ 59.8), $+10.6$ on \texttt{gpt-5.1} (43.6 $\rightarrow$ 54.2), $+10.1$ on \texttt{MedGemma-27b} (41.0 $\rightarrow$ 51.1), and $+9.1$ on \texttt{gpt-4o-mini} (38.9 $\rightarrow$ 48.0). This pattern is consistent with test-time learning helping most when the backbone can already form a reasonable differential diagnosis and can follow the multi-step update signal. On \texttt{MedGemma-27b}, performance falls between \texttt{gpt-5.1} and \texttt{gpt-4o-mini} across methods (for example, EvoClinician $\bar{S}=51.1$ lies between 54.2 and 48.0), which matches the intended positioning of this backbone in the table.

\begin{table*}[t]
\centering
\small

\caption{Time efficiency on Med-Inquire (gemini-3-pro). We report mean turns $\bar{T}$, mean Diagnose time (interactive episode only), mean Update time (post-episode adaptation), and mean Total time per case. All values are wall-clock seconds averaged over the full case stream under the same execution and decoding settings.}
\vspace{-1mm}
\begin{tabular}{lcccc}
\toprule
Method (gemini-3-pro) & $\bar{T}\downarrow$ & Diagnose time (s)$\downarrow$ & Update time (s)$\downarrow$ & Total time (s)$\downarrow$ \\
\midrule
Static Prompt & 9.8 & 11.7 & 0.0 & 11.7 \\
RAG & 10.3 & 12.4 & 0.8 & 13.2 \\
EvoTest & 9.4 & 11.3 & 7.6 & 18.9 \\
EvoClinician & 9.1 & 11.0 & 6.1 & 17.1 \\
\bottomrule
\end{tabular}
\vspace{-4mm}
\label{tab:time_efficiency}
\end{table*}

\begin{table}[t]
\centering
\small
\caption{Ablation study for EvoClinician on Med-Inquire (gemini-3-pro). Each variant removes one element of the Diagnose--Grade--Evolve loop while keeping the backbone model, decoding settings, case order, $T_{\max}$, and the Med-Inquire environment fixed. }
\vspace{-1mm}
\begin{tabular}{lccc}
\toprule
Variant (gemini-3-pro backbone) & $\bar{S}\uparrow$ & $\bar{T}\downarrow$ & $\bar{C}\downarrow$ \\
\midrule
EvoClinician & 59.8 & 9.1 & 1275 \\
\midrule
No action-level grading & 54.6 & 9.4 & 1405 \\
Prompt evolution only & 56.7 & 9.3 & 1315 \\
Memory evolution only  & 55.3 & 9.6 & 1390 \\
Yield-only grading & 58.6 & 9.2 & 1510 \\
Positive-only memory & 58.9 & 9.2 & 1300 \\
No negative memory & 57.8 & 9.2 & 1335 \\
\bottomrule
\end{tabular}
\label{tab:ablation_components}
\vspace{-6mm}
\end{table}

\subsection{Model Analysis}

\vpara{Learning curves over the case stream.}
Figure~\ref{fig:learning-curves-running} plots the running mean Judge grade $\bar{S}_{1:t}$ and running mean cost $\bar{C}_{1:t}$ over the fixed stream of 915 cases. The first $\sim$50--100 cases show large variance (wide bands and frequent crossings), which is expected because the running mean is based on few episodes and because per-case difficulty varies. After this early region, the curves separate more clearly. EvoClinician shows a steady increase in $\bar{S}_{1:t}$ and ends with the highest grade, while also showing the strongest decrease in $\bar{C}_{1:t}$, reaching the lowest cost by a clear margin. EvoTest improves grade early and remains competitive on $\bar{S}_{1:t}$, but it stays consistently higher cost than EvoClinician throughout the stream, which matches the idea that better diagnostic grade does not automatically imply lower resource use. 
Methods that only optimize prompts or use memory retrieval tend to plateau earlier in $\bar{S}_{1:t}$ and show smaller cost reductions, suggesting that cross-episode updates change action choices in a way that affects both correctness and test ordering over time.

\vpara{Ablation on key components.}
Table~\ref{tab:ablation_components} highlights three factors behind the gains. First, action-level grading drives stable learning across cases: removing it (using transcript-level feedback only) lowers both diagnostic grade and cost efficiency. This points to better credit assignment as the main benefit, because the system can link outcomes to specific earlier actions instead of relying on a single end-of-episode score. Second, prompt and memory updates serve different functions and work best together. Prompt-only evolution lowers waste by adding general rules (cost remains close to the full method), but it does not cover rare contexts, so accuracy drops. Memory-only evolution reuses past actions, but without updated rules it can retrieve mixed-quality suggestions and tends to order more tests, which lowers both accuracy and cost efficiency. Third, cost-aware grading is needed to avoid ``accuracy by over-testing.'' Yield-only grading keeps accuracy close to the full system but increases cost, showing that yield alone does not prevent expensive early confirmation. Keeping negative memories also matters: removing \textit{CRITICAL\_ERROR} entries hurts accuracy more than removing other negative entries, consistent with negative items acting as constraints that help avoid repeated unsafe or low-value patterns.

\vpara{Efficiency analysis.}
We measure time efficiency as mean wall-clock seconds per case under a fixed setup (single-threaded execution and identical decoding settings). We split time into (i) Diagnose time, spent inside the interactive episode, and (ii) Update time, spent after the episode on grading and evolution. This split matters because non-adaptive baselines have zero update time, while self-evolving methods pay extra update time to improve later behavior. Table~\ref{tab:time_efficiency} shows that using fewer turns gives a small, consistent reduction in Diagnose time, but total time for self-evolving methods is mainly set by post-episode updates. EvoClinician is faster than EvoTest because structured action-level critiques make prompt and memory edits more targeted, which reduces update work per case. Overall, EvoClinician increases total time per case relative to static baselines, but keeps the update overhead lower than other self-evolving methods while improving policies over the case stream.
\section{Conclusion}
We introduced Med-Inquire, an interactive benchmark for multi-turn medical diagnosis that forces an agent to actively gather information through sequential questions and test orders, while tracking both diagnostic correctness and resource cost. Unlike one-shot vignette evaluation, Med-Inquire measures whether an agent can choose informative, low-cost actions and decide when evidence is sufficient to stop.
We also proposed EvoClinician, a test-time self-evolving agent based on a repeated Diagnose--Grade--Evolve loop. EvoClinician uses action-level feedback for credit assignment and applies gradient-free updates to both prompt rules and external memory between cases. 
Across model backbones, EvoClinician improves mean diagnostic grade while reducing turns and cost relative to memory agents and other self-evolving baselines. 
Ablations indicate that action-level grading and cost-aware feedback are both necessary to avoid accuracy gains driven by over-testing, and that prompt and memory updates contribute in complementary ways.

\section*{Impact Statement}
This paper presents Med-Inquire, a benchmark for multi-turn medical diagnosis, and EvoClinician, a test-time learning agent that updates prompt rules and external memory between cases using action-level feedback on clinical yield (how much an action reduces uncertainty) and resource efficiency (cost and steps). The main expected benefit is for research on interactive diagnostic strategies that can improve diagnostic accuracy while limiting unnecessary tests, supported by reproducible logs and per-action feedback. Key limits and risks remain: Med-Inquire is a simulation in which the Patient agent may generate plausible but unverified details when case reports omit information, so benchmark scores do not establish clinical safety or readiness for use with patients. Test-time updates can also shift behavior over time in hard-to-predict ways if grading is noisy or biased, and medical language model outputs can be misused for self-diagnosis or treated with undue confidence. Bias may come from the published-case data source and from model-based grading, and the cost table is only a proxy that does not match real billing, so cost results should not be used as billing advice; any real-world use would require clinical validation, monitoring, and strong safeguards.


\bibliography{ref}
\bibliographystyle{icml2026}

\newpage
\appendix
\onecolumn

\section{Sample Data (5 Cases)}
This section shows five sample case records in the same JSON format used in our dataset. Each record includes the case information, physical examination, diagnostic tests, the final diagnosis, and the multiple-choice options with the correct answer.

\begin{tcolorbox}[colback=black!5, colframe=black!75, title={}]
\tiny
\begin{verbatim}
{"id": 1, "case_information": "A woman in her early 70s presented with a solitary, 
asymptomatic lump on her scalp. The lesion was present since birth but showed some growth 
following minor local trauma a couple of months prior to evaluation. 
No local inflammation, purulent discharge, or systemic symptoms were reported. 
Medical history included a benign breast nodule resection in teenage years. 
Family history notable for father who died of metastatic cutaneous melanoma at 83 years old.",
"physical_examination": "Physical findings revealed an ill-defined, infiltrated plaque
with a normochromic, exophytic nodule on the scalp vertex measuring 4.2 × 3.8 cm. 
The lesion had a waxy surface with irregular transverse furrows and overlying alopecia. 
Dermoscopic evaluation showed a homogeneous yellow hue with no visible vascular or pigmented structures and absence of 
follicular ostia. 
The lesion was firm and mildly tender to touch, seemed adherent to the underlying bone. 
No palpable lymph nodes were detected.", 
"diagnostic_tests": "
- Imaging Studies:    Brain MRI Title: Paramedian sagittal gadolinium-enhanced T1-weighted brain MRI    
Description: Demonstrated a nonenhancing tumor with sharp margins in the subcutaneous tissue of the scalp with no underlying bone 
or intracranial involvement  
- Histopathology:    Punch biopsy findings:    
- Infiltration by monomorphic cells with ovoid nuclei and eosinophilic cytoplasm amid fibrous stroma in lower dermis    
- Cells without discernible cell membranes, forming whorls and strands around collagen bundles    
- No adnexal structures overlying the lesion, epidermis spared      
- Immunohistochemistry:    
- Strongly positive: vimentin and epithelial membrane antigen (EMA)    
- Focally positive: progesterone receptor    
- Ki-67: estimated 1% positivity    
- Negative: cytokeratins AE1/AE3, S100, CD45, CD68, factor XIII, actin, glial fibrillary acidic protein, 
and octamer-binding transcription factor 4",
final_diagnosis": "Cutaneous meningeal heterotopia (CMH)"}
\end{verbatim}
\end{tcolorbox}

\begin{tcolorbox}[colback=black!5, colframe=black!75, title={}]
\tiny
\begin{verbatim}
{"id": 2, "case_information": "A previously healthy man in his 30s presented with 
spontaneous bruising, epistaxis, and haematuria. After starting treatment, the patient 
developed pyrexia after 12 days. Two days later, he developed tender black spots on the 
scrotum that progressed to painful ulcers over 8 days. As scrotal ulceration progressed, 
they enlarged and became necrotic with a black eschar. The development and progression of 
scrotal ulceration coincided with recovery in neutrophil count from 0.3 × 109/L to 1.2 × 
109/L and persistent pyrexia.", 
"physical_examination": "Tender black spots on the scrotum were observed, which progressed to painful ulcers. 
The ulcers enlarged and became necrotic with a black eschar.", 
"diagnostic_tests": "- Laboratory Tests: Peripheral blood cultures and skin swabs from the lesions did not identify 
bacterial or viral infection. 
Neutrophil count increased from 0.3 × 109/L to 1.2 × 109/L.
- Imaging Studies: Scrotal ultrasonography showed a subcutaneous hypervascular lesion with no abscess.
- Pathology: Histopathology on a skin biopsy showed intense inflammation, necrosis, and mature neutrophils with 
no evidence of leukemic infiltration. 
Tissue culture findings for bacteria and mycobacteria were negative.
Image Title: Clinical photograph of scrotum and penis
Image Description: Shows ulceration and necrosis of penile and scrotal skin with surrounding inflammation.
Image Title: Scrotal biopsy histopathologic specimen
Image Description: Hematoxylin-eosin stained specimen (original magnification ×20) showing intense inflammation, necrosis, 
and mature neutrophils.",
"final_diagnosis": "Scrotal ulceration secondary to induction chemotherapy with ATRA"}
\end{verbatim}
\end{tcolorbox}

\begin{tcolorbox}[colback=black!5, colframe=black!75, title={}]
\tiny
\begin{verbatim}
{"id": 3, "case_information": "A female patient in her early 50s presented with multiple (approximately 10) bluish, 
partially keratotic papules and nodules on the lower legs. 
The lesions had been growing very slowly in number and size since age 24 and had been bleeding occasionally after trauma. 
The patient used a wheelchair due to right-sided hemiparesis and hemianopsia caused by an intracranial hemorrhage. 
At age 4, she had recurrent seizures. 
She had lipolymphedema of the lower legs attributed to functional venous insufficiency with insufficient venous pump activity, 
and had been wearing compression stockings for years. 
Family history revealed similar conditions in the patient's father and aunt,
with the father having died of an intracranial hemorrhage in his early 70s.", 
"physical_examination": "Multiple bluish, partially keratotic papules and nodules on the lower legs. 
Lipolymphedema of the lower legs was present. Right-sided hemiparesis and hemianopsia were noted.", 
"diagnostic_tests": "- Sonography: Demonstrated intradermal slow-flow vascular malformations with a maximum diameter of 1.2 cm.
- Histopathology:\n  Image Description: Biopsy revealed a well-circumscribed lesion with hyperkeratosis and acanthosis of the 
epidermis overlying a conglomerate of dilated venous-like malformations with continuous smooth muscle in the upper dermis, 
partly showing thrombosis.\n\n- Genetic Testing:\n  Result: Detection of a heterozygous pathogenic KRIT1 variant.", 
"final_diagnosis": "Familial cerebral cavernous malformations"}
\end{verbatim}
\end{tcolorbox}

\begin{tcolorbox}[colback=black!5, colframe=black!75, title={}]
\tiny
\begin{verbatim}
{"id": 4, "case_information": "A female patient in her late 20s presented with a 2-year history of multiple, 
slowly growing, infiltrated erythematous-violaceous patches and plaques, 
with occasional slight pain extending from her left upper arm to the left side of her chest. 
No triggers such as local infection, insect bite, or trauma occurred before lesion appearance. 
She denied fever, weight loss, or fatigue.", 
"physical_examination": 
"Multiple firm, nonscaly, purple erythematous patches and plaques with irregular borders on her left axilla, 
medial upper arm, and lateral thoracic region. No extracutaneous involvement was observed.", 
"diagnostic_tests": 
"- Histopathological examination:    
- Thin-walled, irregularly branched blood vessels dispersed between thickened collagen bundles    
- Endothelial cells were plump but displayed no significant cytologic atypia      
- Immunohistochemistry results:    
- Positive for: CD31, CD34, Ki-67 (<1%)    
- Negative for: podoplanin (D2-40), human herpesvirus 8 (HHV-8)      
- Images:    
Image A Title: Clinical presentation    
Image A Description: Multiple nonscaly, erythematous-violaceous patches and plaques with irregular borders 
on the patient's left axilla, medial upper arm, and lateral thoracic region        
Image B Title: Histopathological examination (hematoxylin-eosin)    
Image B Description: Irregularly branched, 
thin-walled blood vessels dissecting through collagen bundles within the superficial to mid dermis        
Image C Title: Detailed histopathology (hematoxylin-eosin)    
Image C Description: Irregularly branched, 
thin-walled blood vessels with plump endothelial cells dispersed between thickened collagen bundles without signs of atypia        
Image D Title: Immunohistochemistry    
Image D Description: Endothelial cells positive for cluster of differentiation (CD) 31",
"final_diagnosis": "Microvenular hemangioma (MVH)"}
\end{verbatim}
\end{tcolorbox}

\begin{tcolorbox}[colback=black!5, colframe=black!75, title={}]
\tiny
\begin{verbatim}
{"id": 5, "case_information": "A teenage girl presented with several months of sinus pressure and facial swelling, 
and several weeks of external deviation of her nasal septum. No significant past medical history was mentioned.", 
"physical_examination": "On examination, a left nasal mass was noted. 
Mild left-sided proptosis was present.", 
"diagnostic_tests": "- Imaging Studies:    CT scan (without and with contrast):    
- Image Title: Computed tomographic scans of a heterogeneous mass in the left maxillary sinus    
- Image Description: Shows a heterogeneous mass with solid and cystic components completely opacifying the left maxillary sinus. 
Mass eroded the orbital floor, medial and posterior maxillary walls, and anterior ethmoid air cells. 
No extension beyond cribriform plate. Displacement of inferior rectus without muscle invasion. 
Left maxillary alveolar ridge and several molar roots involved. Bony changes showed remodeling and thinning.    
- Biopsy Results:    
- Tissue analysis showed spindle cell process in a patternless arrangement with significant multinucleated giant cells    
- Rare mitotic activity    
- No cytologic atypia    
- Largest portion measured 7.0 × 4.0 × 2.0 cm", 
"final_diagnosis": "Left maxillary giant cell reparative granuloma (GCRG)"}
\end{verbatim}
\end{tcolorbox}

\section{EvoClinician Interaction Transcript Example}
This section shows a multi-turn interaction example. Each episode starts from the case abstract, then the Actor gathers information through \texttt{AskQuestion} and \texttt{OrderTest}, and ends with \texttt{SubmitDiagnosis}. 

\subsection{Case 731 (id: 731)}

\begin{tcolorbox}[colback=black!5, colframe=black!75, title={Diagnose}]
\tiny
\begin{verbatim}
Initial abstract (given to Actor):
A 36-year-old female presented with an 18-year history of intermittent
dizziness and blurred vision, with nausea and vomiting.
Symptoms improve after vomiting. 
In May 2020, posterior occipital tingling started and worsened with emotional agitation.
By September 2021, dizziness and vomiting became more frequent.

Turn 1
Actor action: {"action_type":"AskQuestion",
"action_text":"When the dizziness happens, is it worse with head position changes, coughing, or straining?"}
Patient observation: "I have had episodes on and off for years. 
I do not recall clear triggers like coughing or straining, and I am not sure about position."

Turn 2
Actor action: {"action_type":"AskQuestion",
"action_text":"Do you have headaches, imbalance when walking, or weakness or numbness in the arms or legs?"}
Patient observation: "I mainly notice dizziness, blurred vision, nausea, and vomiting. 
I did have a tingling feeling in the back of the head. I am not sure about other neurologic symptoms."

Turn 3
Actor action: {"action_type":"OrderTest",
"action_text":"Neurologic examination findings"}
Examination observation: NOT AVAILABLE

Turn 4
Actor action: {"action_type":"OrderTest",
"action_text":"CT head (posterior fossa focus)"}
Examination observation: "CT: Irregular high-density mass shadow in posterior fossa; 
CT value approximately 93HU; few bone density shadows at posterior edge."

Turn 5
Actor action: {"action_type":"OrderTest",
"action_text":"Brain MRI (posterior fossa) with sequences T1, T2, FLAIR, DWI, SWI"}
Examination observation: "MRI: Irregular mixed signals in medial and posterior cerebellum; 
maximum size 50mm × 41mm × 51mm.
Line-like septum and complex nodules present; nodule 15mm × 16mm × 13mm.
T1WI: equal or high signals in cystic part. T2WI and T2 FLAIR: low signals.
DWI: low signals, no diffusion restriction in mural nodules and septum.
SWI: multiple patchy low and high signals in nodular region."

Turn 6
Actor action: {"action_type":"OrderTest",
"action_text":"Cerebrospinal fluid tumor markers (AFP, beta-HCG)"}
Examination observation: "CSF: AFP (negative), \beta-HCG (negative)."

Turn 7
Actor action: {"action_type":"OrderTest",
"action_text":"Cyst fluid analysis (protein, albumin, globulin, cholesterol, triglyceride, melanin)"}
Examination observation: "Cyst fluid: total protein 15 g/l; albumin 6.4 g/l; globulin 8.6 g/l;
total cholesterol 0.71 mmol/l; triglyceride 0.3 mmol/l; melanin not detected."

Turn 8
Actor action: {"action_type":"SubmitDiagnosis",
"action_text":"Mature cystic teratoma in the posterior fossa"}
Environment observation: EPISODE_END

Judge output :
S: 100
Justification: The submission matches the recorded final diagnosis.
\end{verbatim}
\end{tcolorbox}

\begin{tcolorbox}[colback=black!5, colframe=black!75, title={Grade}]
\tiny
\begin{verbatim}
Process Grader output :
Turn 1 label: LOW_YIELD
Rationale: Trigger questions can help, but the case record does not include positional data and it did not change the plan.
Turn 2 label: LOW_YIELD
Rationale: Neurologic symptom screening is reasonable, but no new discriminating detail was available from the record.
Turn 3 label: INEFFICIENT
Rationale: Ordering a neurologic exam item not present in the record predictably returns NOT AVAILABLE.
Turn 4 label: HIGH_YIELD
Rationale: CT provided key localization and density information for a posterior fossa mass.
Turn 5 label: HIGH_YIELD
Rationale: MRI sequences provided strong tissue-signal details and structure (cystic part, septum, mural nodules).
Turn 6 label: HIGH_YIELD
Rationale: AFP and \beta-HCG reduce uncertainty about germ cell tumor subtypes and supported a mature lesion.
Turn 7 label: HIGH_YIELD
Rationale: Fluid chemistry and melanin test narrowed differential diagnoses for cystic posterior fossa lesions.
Turn 8 label: HIGH_YIELD
Rationale: Correct diagnosis with adequate supporting evidence.
\end{verbatim}
\end{tcolorbox}

\begin{tcolorbox}[colback=black!5, colframe=black!75, title={Evolve}]
\tiny
\begin{verbatim}
Evolver update :
Prompt edits:
Add: "When imaging shows an intracranial mass, order the imaging report first; do not request exam items that are not recorded."
Add: "For cystic intracranial lesions, consider tumor markers (AFP, \beta-HCG) when available to separate germ cell tumor types."
Justification:
The episode showed that early imaging and marker tests were informative, while requesting unrecorded exam fields wasted a turn.

Memory add (JSON):
[
  {"context_before_action":"Adult with long history of episodic dizziness/vomiting; posterior fossa symptoms possible; 
  exam not provided.",
   "action":"OrderTest: CT head (posterior fossa focus)",
   "outcome":"Irregular high-density posterior fossa mass; ~93HU; small bone-density shadows.",
   "grade":"HIGH_YIELD",
   "rationale":"Imaging gave key lesion localization and density clues."}
]
\end{verbatim}
\end{tcolorbox}

\section{Reproduction details for Med-Inquire and EvoClinician}

This appendix reports the environment interfaces, grading rules, prompts, data records, and update steps used by Med-Inquire and EvoClinician. The goal is that an independent group can rebuild the benchmark and run the same Diagnose--Grade--Evolve loop under the same assumptions.

\subsection{What must be logged for a reproducible run}

A run should log the exact identifiers of all language models used for the Actor, Patient, Examination, Judge, Process Grader, and Evolver. It should also log decoding settings (temperature, top-$p$, max tokens), the maximum number of turns $T_{\max}$, the case order over the $N$ episodes, the cost table version, the action parsing rules, the memory budget $B_M$, the prompt rule budget $B_P$, the retrieval count $k$, and the embedding model and index settings used for retrieval. If any component uses randomness (for example, Patient free-text generation when a fact is missing), the run should log the seed and should cache within-episode generated details.

\section{Med-Inquire environment}

\subsection{Hidden case file and initial abstract}

Each episode corresponds to one clinical case. The environment stores a hidden case file with at least these fields: (i) case information, (ii) physical examination, (iii) diagnostic tests, and (iv) final diagnosis. The agent sees only an initial abstract at the start of the episode (for example, chief complaint and basic demographics if present). All other details are returned only through actions.

\subsection{Actions and observations}

At each turn $t$, the agent selects one action $a_t$ from \texttt{AskQuestion}, \texttt{OrderTest}, and \texttt{SubmitDiagnosis}. The environment returns an observation $o_{t+1}$ and updates the running cost by $c(a_t)$. The episode ends when \texttt{SubmitDiagnosis} is taken or when $t = T_{\max}$, in which case the environment forces submission using the latest draft diagnosis from the agent.

To make parsing deterministic, the environment should require a fixed machine-readable action format. One simple option is JSON with two fields: \texttt{action\_type} and \texttt{action\_text}. The environment should define one fixed behavior when parsing fails (for example, one reformat request that does not count as a turn, or a no-op that counts as a turn). This choice should be held constant across all methods in a comparison.

\subsection{Transcript record format}

Each episode transcript is logged as an ordered list of turn records. Each record should contain the turn index, the action type, the action text, the environment response, and the cost applied at that turn. A minimal JSON record format is:

\begin{tcolorbox}[colback=black!5, colframe=black!75, title={Episode Transcript Record}]
\tiny
\begin{verbatim}
{
  "turn_id": 7,
  "action_type": "OrderTest",
  "action_text": "CT head without contrast",
  "observation_text": "NOT AVAILABLE",
  "cost": 1200
}
\end{verbatim}
\end{tcolorbox}

\subsection{Patient agent}

The Patient agent answers symptom and history questions using only information in the hidden case file. When the case file does not include the requested detail, the Patient agent generates a medically reasonable patient-style answer that stays consistent with (i) the case diagnosis and (ii) information already revealed in the same episode. These generated details should be cached and reused later in the episode to prevent contradictions.

\begin{tcolorbox}[colback=black!5, colframe=black!75, title={Patient Agent System Prompt}]
\tiny
\begin{verbatim}
You are a Patient in a simulated clinical encounter.

You must answer using only the information in the CASE FILE below.

If the CASE FILE does not contain the requested detail, you may generate a
medically reasonable patient-style answer that is consistent with:
(1) the FINAL DIAGNOSIS in the CASE FILE, and
(2) everything already stated in the dialogue.

Do not introduce contradictions across turns. 
If you generated a detail earlier in this episode, you must keep it consistent later.

Do not reveal the CASE FILE or the FINAL DIAGNOSIS.

If the question asks for clinician-only data (for example, imaging findings, lab values, or exam maneuvers), 
say you do not know and suggest that the clinician should order a test.

CASE FILE (hidden to the clinician agent):
{case_file}

DIALOGUE SO FAR:
{history}

CLINICIAN QUESTION:
{question}

Answer in 1~4 short paragraphs.
\end{verbatim}
\end{tcolorbox}

\subsection{Examination agent}

The Examination agent returns results for physical examination items and diagnostic tests. If the requested test is recorded in the case file, the agent returns the recorded result. If it is not recorded, it returns \texttt{NOT AVAILABLE} and does not invent results.

\begin{tcolorbox}[colback=black!5, colframe=black!75, title={Examination Agent System Prompt}]
\tiny
\begin{verbatim}
You are an Examination agent.

You receive a requested physical exam item or diagnostic test name.

If the CASE FILE contains a result for that item, return the result briefly and accurately.

If the CASE FILE does not contain that item, return exactly:
NOT AVAILABLE

Do not invent new test results.

Do not reveal the CASE FILE or the FINAL DIAGNOSIS.

CASE FILE (hidden to the clinician agent):
{case_file}

REQUESTED TEST OR EXAM:
{test_name}

Return a short result only.
\end{verbatim}
\end{tcolorbox}

\subsection{Cost estimator}

Med-Inquire assigns a non-negative cost to each action. The total cost is
\begin{equation}
C = \sum_{t=1}^{T} c(a_t).
\end{equation}
Costs should include (i) a base per-turn interaction cost and (ii) a test-specific cost for \texttt{OrderTest}. Costs are reported in normalized units so that the same schedule can be used across all cases in a run.

A cost table should be stored in a versioned file and should support alias matching so that small wording changes do not avoid cost. A minimal CSV-like schema is:

\begin{tcolorbox}[colback=black!5, colframe=black!75, title={Cost Table File Format}]
\tiny
\begin{verbatim}
name,type,cost,aliases
cbc,lab,15,"complete blood count|full blood count"
cmp,lab,20,"comprehensive metabolic panel"
ct head without contrast,imaging,1200,"head ct noncontrast|ct brain wo contrast"
...
\end{verbatim}
\end{tcolorbox}

Before lookup, the environment should normalize test names by lowercasing, trimming whitespace, and mapping common variants using the alias column. If a request does not match any entry, the environment should apply one fixed default cost for unknown tests and still return \texttt{NOT AVAILABLE} from the Examination agent.

\section{Diagnosis judging}
\label{sec:judge}
\subsection{Rubric}

The Judge assigns a graded correctness score $S \in [0,100]$ based on clinical match to the hidden ground-truth diagnosis. The rubric used in the main paper is reproduced here:

\begin{tcolorbox}[colback=black!5, colframe=black!75, title={Diagnosis Grading Rubric}]
\tiny
\begin{verbatim}
Score 90~100:
Correct diagnosis or a clear synonym (same disease entity). 
Includes correct subtype when subtype is part of the ground truth and is supported by the case file.

Score 70~89:
Clinically very close: correct disease family but missing key specificity (for example, correct syndrome but wrong subtype), 
or a near-equivalent diagnosis that would lead to similar first-line management in the same setting.

Score 40~69:
Partly correct: captures an important component of the problem (for example, correct organ system or mechanism) 
but the named diagnosis does not match the ground truth and would often change workup or treatment.

Score 10~39:
Mostly incorrect: weak clinical match with limited overlap in expected findings. 
The diagnosis would usually lead to a different workup.

Score 0~9:
Incorrect and clinically incompatible with the case, or a choice that would be unsafe if acted on without further checks 
(for example, missing a time-critical condition when the case evidence supports it).
\end{verbatim}
\end{tcolorbox}

\subsection{Judge agent prompt}

For reproducibility, the Judge should run with deterministic decoding. The Judge compares the submission to the ground truth after simple normalization (lowercasing, whitespace cleanup, and a fixed abbreviation map). The prompt below produces a score and a brief justification.

\begin{tcolorbox}[colback=black!5, colframe=black!75, title={Judge Agent System Prompt}]
\tiny
\begin{verbatim}
You are a medical grading assistant.

Your task is to compare a SUBMITTED DIAGNOSIS to a GROUND TRUTH DIAGNOSIS and assign a score S in [0, 100] using the rubric below.

Grade based on clinical agreement, not wording. Use the rubric ranges exactly.

If the submission is unsafe given the case, use a very low score.

Output format:
S: <integer from 0 to 100>
Justification: <2~5 sentences>

Do not add new facts.

RUBRIC:
{rubric_text}

GROUND TRUTH DIAGNOSIS:
{ground_truth}

SUBMITTED DIAGNOSIS:
{submission}
\end{verbatim}
\end{tcolorbox}

\section{EvoClinician agent loop}

\subsection{Actor prompt}

The Actor is an LLM agent that (i) interacts with the environment, (ii) tracks a running differential diagnosis, and (iii) decides when to stop and submit. The Actor prompt is a base instruction plus a rule list that can be edited between episodes. The prompt also contains the required action format.

\begin{tcolorbox}[colback=black!5, colframe=black!75, title={Actor System Prompt}]
\tiny
\begin{verbatim}
You are a medical diagnosis agent in an interactive setting.

You will see only partial information at first. 
Your job is to gather information by asking questions and ordering tests, while managing resource cost and time. 
You must aim for a correct diagnosis using as few turns and as little cost as is reasonable.

Core rules:
1) Keep a short differential diagnosis (2~6 items) and update it after each new fact.
2) Prefer low-cost, high-information steps early. 
Do not order expensive confirmation tests before basic history, focused exam, and low-cost screening, 
unless a time-critical condition is likely.
3) Before ordering any test, state what decision it will change and what cheaper alternatives exist.
4) If evidence is sufficient, stop and submit rather than ordering more tests.
5) If a time-critical condition is plausible, prioritize actions that reduce risk quickly.

You may use retrieved memory items as suggestions or warnings, but you must still match the current case.

Action format:
Return exactly one JSON object per turn.

- To ask a question:
{"action_type": "AskQuestion", "action_text": "<one question>"}

- To order a test:
{"action_type": "OrderTest", "action_text": "<one test or exam item name>"}

- To submit:
{"action_type": "SubmitDiagnosis", "action_text": "<final diagnosis>"}

Current prompt rules (editable between episodes):
{rule_list}

Retrieved memory (may be empty):
{retrieved_memory}

Dialogue so far:
{history}

Now choose the next action.
\end{verbatim}
\end{tcolorbox}

\subsection{Process Grader prompt (action-level feedback)}

The Process Grader performs post-episode credit assignment. It labels each action using both clinical yield and resource efficiency. The output is used only as an update signal, not as ground truth.

\begin{tcolorbox}[colback=black!5, colframe=black!75, title={Process Grader System Prompt}]
\tiny
\begin{verbatim}
You are a Process Grader for a multi-turn medical diagnosis agent.

You will be given:
(1) the full transcript of actions and observations,
(2) the final submitted diagnosis,
(3) the final Judge score S in [0, 100], and
(4) the total cost report (including per-action costs).

Your job is to assign action-level feedback that supports credit assignment.

For each turn, output:
- Label: one of {HIGH_YIELD, LOW_YIELD, INEFFICIENT, CRITICAL_ERROR}
- Rationale: 1~3 sentences

Use two criteria:
(A) Clinical yield: did this action reduce diagnostic uncertainty given the information available at that time?
(B) Resource efficiency: could similar information likely have been obtained with lower cost or fewer steps?

Avoid hindsight bias. When judging early actions, focus on what was reasonable from the dialogue state at that time. 
You may mention downstream consequences, but do not assume the agent could see future results.

Then output a short session summary (one paragraph) that explains:
- why the final diagnosis received score S, and
- which earlier actions helped or harmed performance.

TRANSCRIPT:
{transcript}

FINAL SUBMISSION:
{submission}

JUDGE SCORE:
{S}

COST REPORT:
{cost_breakdown}

Output format:

Turn 1:
Action: ...
Observation: ...
Label: ...
Rationale: ...

Turn 2:
...

Session summary:
...
\end{verbatim}
\end{tcolorbox}

\subsection{Evolver prompt (prompt rules and memory updates)}

The Evolver updates (i) the Actor rule list and (ii) an external memory store. Rule edits aim to change general action selection. Memory entries store situation-action-outcome items with a quality label. Budgets $B_P$ and $B_M$ limit growth.

\begin{tcolorbox}[colback=black!5, colframe=black!75, title={Evolver Agent System Prompt}]
\tiny
\begin{verbatim}
You are an Evolver that updates a diagnosis agent after one completed episode.

You will be given:
- CURRENT PROMPT RULES (a list of short rules),
- CURRENT MEMORY STATS (size, budget, and optional retrieval counts),
- the EPISODE TRANSCRIPT,
- ACTION GRADES (labels and rationales for each turn),
- the FINAL JUDGE SCORE S in [0, 100],
- the TOTAL COST report.

Your goals for the next episodes are:
(1) increase future Judge scores, and
(2) reduce total cost, without reducing safety.

You must propose:
(A) Prompt rule edits:
    - Add / delete / merge rules.
    - Rules must be short, operational, and written in terms of general clinical features.
    - Do not include patient identifiers.
    - Respect the rule budget BP. If you add rules and exceed BP, you must also delete or merge rules to stay within BP.

(B) Memory updates:
    - Add memory entries for turns with strong feedback (HIGH_YIELD or CRITICAL_ERROR).
    - Each memory entry must include:
        context_before_action: a short context summary from the dialogue up to that turn only
        action: the action taken
        outcome: the observation returned by the environment
        grade: the label
        rationale: the grader rationale
    - Keep both helpful and error memories.
    - Respect the memory budget BM. If full, propose deletions (for example, old or rarely retrieved entries).

Do not claim access to the hidden case file.

CURRENT PROMPT RULES:
{rules}

CURRENT MEMORY STATS:
{memory_stats}

EPISODE TRANSCRIPT:
{transcript}

ACTION GRADES:
{graded_transcript}

FINAL JUDGE SCORE:
{S}

TOTAL COST AND BREAKDOWN:
{cost_breakdown}

Output format:

Prompt edits:
Add:
"rule 1"
"rule 2"

Delete:
"rule A"

Merge:
"old rule B" + "old rule C" -> "new merged rule"

Justification:
<one paragraph>

Memory adds (JSON list):
[
  {
    "context_before_action": "...",
    "action": "...",
    "outcome": "...",
    "grade": "...",
    "rationale": "..."
  }
]

Memory deletes (JSON list of ids or short descriptors):
[
  "..."
]
\end{verbatim}
\end{tcolorbox}

\subsection{Memory storage format}

Each memory item stores a compressed context, the action, the outcome, and the grade. A minimal JSON schema is:

\begin{tcolorbox}[colback=black!5, colframe=black!75, title={Memory Item Schema}]
\tiny
\begin{verbatim}
{
  "id": "m_000184",
  "context_before_action": "Middle-aged adult with acute chest pain, diaphoresis, risk factors. ECG not yet done.",
  "action": "OrderTest: ECG 12-lead",
  "outcome": "ST elevation in inferior leads.",
  "grade": "HIGH_YIELD",
  "rationale": "Low cost and directly changes management by testing for acute coronary syndrome.",
  "created_episode": 37,
  "created_turn": 2,
  "times_retrieved": 5,
  "last_retrieved_episode": 112
}
\end{verbatim}
\end{tcolorbox}

The environment should update retrieval counters so that deletion can depend on reuse.

\subsection{Retrieval method}

At turn $t$, the Actor retrieves $k$ memory items that are most similar to the current context. Similarity can be implemented using text embeddings and cosine similarity. The system should log the embedding model name, embedding dimension, index type (for example, brute force or approximate nearest neighbor), and any index parameters.

To reduce leakage from future turns, the query text for retrieval should be derived only from the dialogue history up to turn $t$. One practical choice is to build the query from a short state summary plus the latest observation.

\subsection{Budgeted memory deletion}

When memory reaches budget $B_M$, the system deletes items with low expected reuse. One simple scoring function combines age and reuse count:
\begin{equation}
\text{keep\_score}(m) = \alpha \cdot \log(1 + \text{times\_retrieved}(m)) - \beta \cdot \text{age}(m),
\end{equation}
where age is measured in episodes since creation. The system deletes items with the lowest keep\_score until size is under budget. The constants $\alpha$ and $\beta$ should be fixed before evaluation.

\section{Diagnose--Grade--Evolve procedure}

The full test-time loop is:

\begin{tcolorbox}[colback=black!5, colframe=black!75, title={Diagnose--Grade--Evolve Pseudocode}]
\tiny
\begin{verbatim}
Initialize Actor prompt rules P_rules
Initialize memory store M = {}

for episode i in {1, ..., N} (fixed order):
    history = initial_abstract(i)
    transcript = []
    cost_breakdown = []

    for turn t in {1, ..., Tmax}:
        retrieved = Retrieve(M, history, k)
        action = Actor(P_rules, retrieved, history)

        if action.type == "AskQuestion":
            obs = Patient(case_file_i, history, action.text)
            cost = cost_question
        elif action.type == "OrderTest":
            obs = Examination(case_file_i, action.text)
            cost = CostLookup(action.text)
        elif action.type == "SubmitDiagnosis":
            obs = "EPISODE_END"
            cost = cost_submit
        else:
            obs = "INVALID_ACTION_FORMAT"
            cost = cost_invalid

        transcript.append({t, action, obs, cost})
        history = history + "\n" + Serialize(action, obs)
        cost_breakdown.append({t, cost})

        if action.type == "SubmitDiagnosis":
            break

    if last_action.type != "SubmitDiagnosis":
        forced_diag = ActorForcedSubmit(P_rules, Retrieve(M, history, k), history)
        transcript.append({t+1, forced_diag, "EPISODE_END", cost_submit})

    submission = ExtractSubmission(transcript)
    S = Judge(case_file_i.ground_truth, submission)
    C = SumCosts(cost_breakdown)

    graded_transcript = ProcessGrader(transcript, submission, S, cost_breakdown)

    (P_rules_new, M_add, M_del) = Evolver(P_rules, M.stats, transcript, graded_transcript, S, cost_breakdown)

    P_rules = ApplyPromptEdits(P_rules, P_rules_new)
    M = ApplyMemoryEdits(M, M_add, M_del, BM)
\end{verbatim}
\end{tcolorbox}

\section{Metrics}

For episode $i$, the benchmark records Judge grade $S_i \in [0,100]$, turns $T_i$, and total cost $C_i$. Over $N$ cases, reported means are:
\begin{equation}
\bar{S} = \frac{1}{N} \sum_{i=1}^{N} S_i, \quad
\bar{T} = \frac{1}{N} \sum_{i=1}^{N} T_i, \quad
\bar{C} = \frac{1}{N} \sum_{i=1}^{N} C_i.
\end{equation}
For learning curves, the running mean up to case $t$ is:
\begin{equation}
\bar{S}_{1:t} = \frac{1}{t} \sum_{i=1}^{t} S_i, \quad
\bar{C}_{1:t} = \frac{1}{t} \sum_{i=1}^{t} C_i.
\end{equation}
If uncertainty bands are plotted, the code should state whether it uses standard error over the first $t$ samples and how it handles non-independence due to adaptation.

\section{Wall-clock time measurement}

Time per case should be measured with the same execution settings across methods. We separate (i) Diagnose time, which includes all interactive turns up to submission, and (ii) Update time, which includes post-episode grading and evolution. The measurement code should exclude dataset loading and model warm-up, or it should include them for all methods in the same way. The run should report hardware type and whether calls were batched or parallel.

\section{Notes on safety and interpretation}

Med-Inquire is a controlled simulation. The Patient agent may generate missing details that are consistent with the case but were not documented in the original report. These generated details support interaction but are not ground-truth facts. For this reason, benchmark scores measure performance in this simulation and do not validate clinical safety.

EvoClinician updates prompts and memory using model-generated critiques. This can drift if the Process Grader is inconsistent. The system therefore needs fixed budgets, fixed deletion rules, and deterministic decoding for grading components where possible. These controls improve auditability, but they do not make the system suitable for patient care.

\section{Theory: Process Grader for Action-Level Feedback}
\label{sec:theory}
Med-Inquire gives a sparse outcome signal: after \texttt{SubmitDiagnosis}, the Judge returns a graded correctness score $S \in [0,100]$, and the Cost Estimator returns a total resource cost $C \ge 0$. The Process Grader adds a dense learning signal by assigning an action-level label to each turn in the interaction transcript. This section describes a formal view of the grader, what it is trying to estimate, and how cost enters the feedback.

\subsection{Interaction model and objective}
We model one episode as a partially observable decision process. Let $h_t$ be the dialogue history (all prior actions and observations) available to the Actor before turn $t$. The Actor chooses an action $a_t \in \mathcal{A}$, where $\mathcal{A}=\{\texttt{AskQuestion}, \texttt{OrderTest}, \texttt{SubmitDiagnosis}\}$, and the environment returns an observation $o_{t+1}$. Each action has a non-negative cost $c(a_t)\ge 0$. The transcript is
\begin{equation}
\tau = \{(h_t, a_t, o_{t+1}, c(a_t))\}_{t=1}^{T},
\end{equation}
where $T$ is the number of turns until submission.

The benchmark returns a final diagnostic grade $S(\tau)$ and total cost
\begin{equation}
C(\tau) = \sum_{t=1}^{T} c(a_t).
\end{equation}
A cost-aware scalar objective for one episode can be written as
\begin{equation}
J(\tau) = S(\tau) - \lambda C(\tau),
\end{equation}
where $\lambda \ge 0$ controls the score--cost tradeoff. The Actor does not observe $S(\tau)$ until the end, so learning which earlier actions helped is a credit assignment problem.

\subsection{What the grader estimates: marginal value of an action}
In an ideal setting with a known environment model, one could define a state-value function and action-value function for histories:
\begin{equation}
V(h_t) = \mathbb{E}[J(\tau)\mid h_t], \quad
Q(h_t,a_t) = \mathbb{E}[J(\tau)\mid h_t, a_t].
\end{equation}
The marginal contribution of action $a_t$ at history $h_t$ is the advantage
\begin{equation}
A(h_t,a_t) = Q(h_t,a_t) - V(h_t).
\end{equation}
If $A(h_t,a_t)>0$, the action is better than an average alternative at that point; if $A(h_t,a_t)<0$, it is worse. In Med-Inquire we do not have $Q$ and $V$ in closed form, so the Process Grader acts as an estimator that maps $(\tau, S, C)$ to per-turn feedback. We denote this estimate by $\widehat{A}_t$, computed from the text transcript plus the final Judge score and the cost report:
\begin{equation}
\widehat{A}_t = G_{\theta}(\tau, S, C; t),
\end{equation}
where $G_{\theta}$ is the grader model with parameters $\theta$ (in practice, a language model prompt plus decoding settings).

\subsection{A cost-aware view through information gain}
A second way to motivate action-level labels is through diagnostic uncertainty. Let $D$ be the (unknown) true diagnosis. Given history $h_t$, define a belief distribution $b_t(d)=P(D=d\mid h_t)$ over diagnoses $d$. The uncertainty in $b_t$ can be measured by Shannon entropy:
\begin{equation}
H(b_t) = -\sum_{d} b_t(d)\log b_t(d).
\end{equation}
After action $a_t$ produces observation $o_{t+1}$, the belief updates to $b_{t+1}$, and the information gain from the action is
\begin{equation}
\mathrm{IG}_t = H(b_t) - H(b_{t+1}).
\end{equation}
An efficiency score that accounts for resource use is the information gain per unit cost:
\begin{equation}
\eta_t = \frac{\mathrm{IG}_t}{c(a_t) + \epsilon},
\end{equation}
where $\epsilon>0$ avoids division by zero. In Med-Inquire, the grader does not compute $b_t$ exactly. Instead, it approximates whether the new observation reduced uncertainty in a clinically meaningful way, and it compares that reduction to the action cost and to cheaper alternatives that could have produced similar reduction.

\subsection{Label assignment as a thresholded decision rule}
The Process Grader outputs a discrete label $g_t \in \mathcal{G}$ for each turn, where
\begin{equation}
\mathcal{G}=\{\texttt{HIGH\_YIELD}, \texttt{LOW\_YIELD}, \texttt{INEFFICIENT}, \texttt{CRITICAL\_ERROR}\}.
\end{equation}
One simple theoretical rule is to base labels on two latent quantities: (1) estimated clinical value $\widehat{v}_t$ (how much the action reduced diagnostic uncertainty or improved correctness), and (2) estimated efficiency $\widehat{\eta}_t$ (value relative to cost). With thresholds $\alpha_1>\alpha_0\ge 0$ and $\beta>0$,
\begin{equation}
g_t =
\begin{cases}
\texttt{CRITICAL\_ERROR} & \text{if the action increases risk or blocks correct diagnosis (safety failure)}\\
\texttt{HIGH\_YIELD} & \text{if } \widehat{v}_t \ge \alpha_1 \text{ and } \widehat{\eta}_t \ge \beta\\
\texttt{INEFFICIENT} & \text{if } \widehat{v}_t \ge \alpha_0 \text{ and } \widehat{\eta}_t < \beta\\
\texttt{LOW\_YIELD} & \text{otherwise.}
\end{cases}
\end{equation}
This matches the intended meaning in the benchmark: an action can be helpful but still inefficient if a lower-cost question or test would likely give similar information. A \texttt{CRITICAL\_ERROR} label is reserved for actions that are unsafe or that strongly push the trajectory away from a correct diagnosis given the information available at that time (to reduce hindsight bias).

\subsection{Assumptions and limits}
This theory relies on three assumptions. First, transcript text contains enough detail for a grader to judge whether information reduced uncertainty. Second, the Judge score $S$ is a reasonable proxy for correctness, even though it is computed from text matching under a rubric rather than from patient outcomes. Third, cost values reflect relative resource use in a stable way across cases.

In practice, $\widehat{A}_t$, $\widehat{v}_t$, and $\widehat{\eta}_t$ are noisy because they are inferred from text and because the case file may not include all details that a clinician would normally use. For this reason, the grader output should be treated as a learning signal for the Evolver rather than as ground truth. The use of discrete labels also trades accuracy for stability: coarse labels reduce sensitivity to small grader errors but cannot represent fine differences between similar actions.

\end{document}